%% file: dsaa19.tex
\documentclass[conference,compsoc]{IEEEtran}

\input{00_head}

\begin{document}
%
\title{Topology-based Clusterwise Regression for User Segmentation and Demand Forecasting}

\author{
\IEEEauthorblockN{
Rodrigo Rivera-Castro \IEEEauthorrefmark{1},
Aleksandr Pletnev \IEEEauthorrefmark{1},
Polina Pilyugina \IEEEauthorrefmark{1},
}
\IEEEauthorblockN{
Grecia Diaz \IEEEauthorrefmark{1}, 
Ivan Nazarov \IEEEauthorrefmark{1},
Wanyi Zhu \IEEEauthorrefmark{2},
Evgeny Burnaev \IEEEauthorrefmark{1}
}
\IEEEauthorblockA{\IEEEauthorrefmark{1}Skolkovo Institute of Science and Technology, \IEEEauthorrefmark{2}Alibaba Cloud Intelligence Business Group \\ Email: rodrigo.riveracastro@skoltech.ru}

}


%


\maketitle

\begin{abstract}
Topological Data Analysis (TDA) is a recent approach to analyze data sets from the perspective of their topological structure. Its use for time series data has been limited. In this work, a system developed for a leading provider of cloud computing combining both user segmentation and demand forecasting is presented. It consists of a TDA-based clustering method for time series inspired by a popular managerial framework for customer segmentation and extended to the case of clusterwise regression using matrix factorization methods to forecast demand. Increasing customer loyalty and producing accurate forecasts remain active topics of discussion both for researchers and managers. Using a public and a novel proprietary data set of commercial data, this research shows that the proposed system enables analysts to both cluster their user base and plan demand at a granular level with significantly higher accuracy than a state of the art baseline. This work thus seeks to introduce TDA-based clustering of time series and clusterwise regression with matrix factorization methods as viable tools for the practitioner.
\end{abstract}


%
\IEEEpeerreviewmaketitle

\section{Originality and Value}\label{sec:originality}
This research presents a machine learning based system developed for a leading provider of cloud computing capable of clustering its customer base into meaningful segments and forecasting demand at a user level. Validated with real data, the approach has yet to be deployed in production. The contributions cover the areas of customer relationship management (CRM) and demand forecasting. The proposed system is ideal for individuals and organizations with domain knowledge but limited understanding of machine learning methods. The contributions are the following:
\begin{enumerate}
\item An industry case of customer base analysis and demand prediction for a major provider of cloud computing.
\item An application of time series clustering using Topological Data Analysis for commercial data.
\item A presentation of two novel cluster ensemble methods.
\item A novel clusterwise regression method using matrix factorization techniques.
\item For reproducibility purposes, an implementation and data set available for download \footnote{\url{https://github.com/rodrigorivera/dsaa19}}.
\end{enumerate}

\section{Problem Statement}
\begin{figure}[!htb]
\begin{center}
\includegraphics[width=\columnwidth]{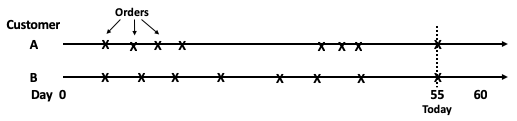} 
\end{center}
\caption{Two customers can share the same Recency, Frequency and Monetary scores. Yet, customer A is likely more alive than B}
\label{fig:general:customer_timelines}
\end{figure}

One of the world's largest cloud computing providers requires a better understanding of its customer base to be able to offer tailored promotions and to assess better the expected future demand for its services. The individual customer demand amounts to millions of time series data to predict. Given the novelty of the cloud computing offerings and the flexibility it offers to customers, historic data is limited, seasonality hard to detect and historic records often non-representative. In summary, the data available at a customer level is limited and hard to work with. As a consequence, traditional forecasting techniques are largely ineffective. Further, popular heuristics for customer segmentation such as the Recency, Frequency, Monetary framework can be misleading with two customers sharing the same score while being very different, as seen in \autoref{fig:general:customer_timelines}.

\section{Introduction}\label{sec:introduction}
\begin{figure*}[!htb]
\begin{center}
\includegraphics[width=\columnwidth*2]{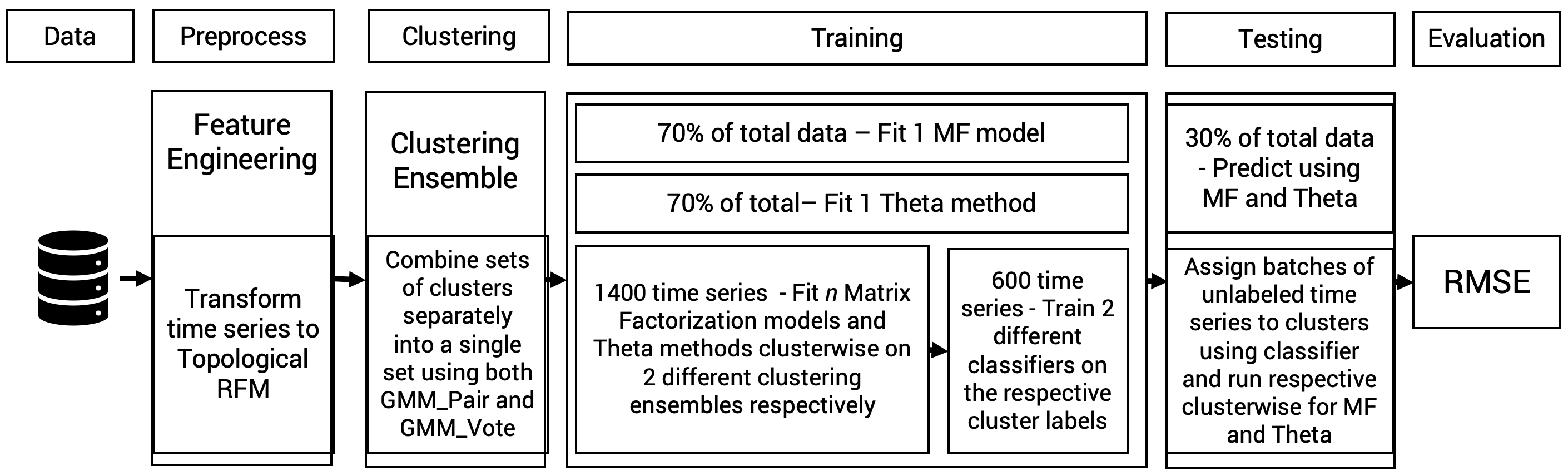}
\end{center}
\caption{Proposed machine learning pipeline combining the Topological RFM framework, \cite{Rivera-Castro2019-qk}, clustering ensemble and clusterwise regression with matrix factorization methods}
\label{fig:general:process_diagram_1}
\end{figure*}

Developing systematic customer insights from customer loyalty data requires a system that can be easily adopted by business analysts in the marketing and demand planning functions of a company. The aim of this work is to present a new time series clustering model combining well-understood managerial heuristics together with predictive methods from the time series clustering and topological data analysis literature such as clusterwise regression. Rather than relying purely on traditional heuristics such as the Recency, Frequency, Monetary (RFM) framework to understand the customer base, this work proposes a machine learning pipeline consisting of multiple stages depicted in \autoref{fig:general:process_diagram_1}. The viability of this system is validated with customer data gathered from two relevant data sets, one proprietary and another public. Similarly, this proposal is not only useful to assist in the managerial decision-making but also to help improve prediction accuracy at a user level.

This work seeks as a first step to introduce Customer Loyalty, the Recency Frequency Monetary, Time Series Clustering and Topological Data Analysis in the Literature Review. It proceeds to present the various elements of the proposed system such as forecasting using Matrix Factorization, clusterwise regression, clustering ensemble and topological RFM. It then presents two relevant data sets, one commonly used in the CRM literature and another novel one from the cloud computing sector. In \autoref{sec:experiments}, a baseline and the methodology for evaluation is introduced. This research concludes with a Discussion and Advice for the Practitioner sections respectively.

\section{Research Abstract and Goals} \label{sec:research_abstract}

\subsection{Aims and Backgrounds}
The objective of this research is to present a system combining customer segmentation inspired by an established managerial framework together with a predictive method that can be accessible to non-technical business experts. CRM models emphasizing timing patters to predict future purchase activities have been proposed previously by \cite{Zhang2015-wa} and \cite{Platzer2016-si} and serve as an inspiration for this work. The use of novel machine learning methods is a promising area with little academic research and insufficient efforts to expose practitioners to them according to \cite{2017arXiv170905548R} and \cite{Rivera_2018}. In addition, over 40\% of analysts still use primarily traditional forecasting methods, \cite{Rivera-Castro2019-dj}. 

\subsection{Significance}
There are significant incentives to develop methods that can be easily adopted by quantitative marketeers. In forecasting, for discrepancies as low as 2\%, it is worth improving the accuracy of a forecast, \cite{Fleisch2003}. Yet, companies struggle hiring the adequate personnel to address these tasks. For example, a survey by \cite{2018esade} reported that over 70\% of surveyed businesses in Europe struggled hiring data science personnel and over 60\% are resorting to internal training to upgrade the skills of existing business analysts. Similarly, by 2020, Vietnam is expected to face a shortage of over 500 000 employees with data science and analytics skills and over 80\% of the local workforce do not have the necessary skill set to fill this gap, \cite{2017apec}. This work seeks to alleviate this situation by presenting a system based on state-of-the-art methods that is both accurate as well as easy to communicate to decision-makers.

\subsection{Research Questions \& Objectives}
The research goal of this work is to propose an approach combining user segmentation and demand forecasting that can be adopted by business practitioners. For this purpose, the study poses the questions: 
1) Although RFM-inspired models are attractive due to their increased accuracy, how can they be extended to also offer forecasting at a granular level? 
2) How can methods based on clustering be integrated to combine both segmentation and prediction? 
To achieve the research goal, two objectives have been assigned: 
a) To review the existing theory on enriching RFM to measure customer loyalty more accurately; 
b) To make a performance comparison between a state of the art baseline and the proposed system.
The object of research is the balance between accessibility and precision of methods for customer segmentation and demand prediction using time series clustering topological data analysis and clusterwise regression within the industry. 
The subject of the research is customer segmentation combined with prediction of customer's next action.
\section{Literature Review} \label{sec:literature_review}
\subsection{Customer Loyalty}
The literature covering Customer Loyalty is vast and a thorough review is out of the scope of this work. This study narrows it down by focusing on the combination of existing frameworks for customer classification extended with machine learning methods. Examples of this are the combination of the Recency Frequency Monetary framework with other techniques and extending its scope can be seen on the recent work of \cite{Zaki2016-oj}. They combined the Net Promoter Score, a survey-based metric commonly used to predict customer satisfaction and repurchase intention, together with RFM; thus, giving additional meaning to NPS by adding a quantitative factor based on purchase history. This study follows the argument made by \cite{Wubben2008-yf} that the use of big data techniques must be used to update the customer loyalty measurement in organizations. Firms benefit from the use of sophisticated and advanced approaches. They help uncover patterns in customer data, which can be linked to business results, \cite{doi:10.1108/JOSM-01-2013-0018}. \cite{Zaki2016-oj} advocate for the use of data mining approaches to evaluate customer loyalty instead of relying solely on survey-based measurement and statistical techniques.
An example of the use of data mining techniques to assess customer loyalty is the work of \cite{hosseini2010}, where they combined RFM and K-Means, a clustering algorithm, to measure the degree of customer loyalty to maximize the profits of B2B businesses. The results showed a significant improvement in the accuracy of the measurement of customer loyalty. Another example of measuring customer loyalty with big data is \cite{tirunillai2014}, where they combined user-generated content from online chats. 
Machine learning based techniques have also found their way in the development of predictors of customer loyalty. An example of this is \cite{doi:10.1108/08876041211223951} for the retail banking sector. Predictors related to demographic factors and customer's perceptions of market conditions were proposed. Other entries in the literature such as \cite{Wubben2008-mc} and \cite{Tamaddoni2016-eo} sought to identify customer churn and compared multiple machine learning classifiers such as Support Vector Machine, logistic regression, boosting, the Pareto/NBD model, a probabilistic model, and the Recency Frequency Monetary model. They concluded that for small data sets, Pareto/NBD offers the best results. Yet, for companies with large customer bases, boosting techniques are a better option. Boosting methods for customer behavior are also discussed by \cite{Martinez2018-us} for non-contractual settings, where they developed customer relevant features and evaluated them using three regressors: Lasso, extreme learning machine and gradient tree boosting.

\subsection{Recency Frequency Monetary}
Recency Frequency Monetary (RFM) is a managerial metric originated in database marketing, a form of direct marketing. In its original form, it seeks to increase response rates by classifying customers into five equal groups based on aspects of their past behavior. As a result, a three-digit number is obtained. The lower the number, the higher the probability of customer churn, \cite{doi:10.1177/1094670506293810}. 
\cite{Blattberg2008-jv} adopted RFM for direct database marketing purposes. Since then, RFM has been found to be an important predictor of future customer life value (CLV) and customer behavior and churn, \cite{Tamaddoni2016-eo, Ballings2012-ey}. \cite{Coussement2008-hw} observed that recency has the highest predictive influence. It has therefore been widely used as a measure of behavioral loyalty since its inception. The model transforms customer transactional data into profitability scores; thus, facilitating the categorization of customers based on their purchasing behavior. To perform an RFM analysis, customers' purchasing patterns must be observed over a predefined period. In its original form, customers are first sorted based on their recency values and then the customer base is divided into five groups. Each customer then receives a rank. The lowest recency value corresponds to a score of 5, whereas the highest obtains a score of 1. Similarly, this procedure is carried out for both frequency and monetary. Thus, each customer obtain a three-digit score corresponding to the combination of each RFM variable rank.

\subsection{Time Series Clustering}
Clustering time-series data is a technique used in many areas to discover patterns. Broadly, clustering represents partition $n$ observations into $k$ clusters, where a cluster is characterized with the notions of homogeneity, the similarity of observations within a cluster, and separation, which is the dissimilarity of observations from different clusters. Formally, this is defined as a set of $n$ observations $X = \{\vec{x}_1,\ldots,\vec{x}_n\}$, where $\vec{x}_i\in \mathbb{R}^m$, and the number of clusters $k < n$, the objective is to partition $X$ into $k$ pairwise-disjoint clusters $P = \{p_1,\ldots,p_k\}$, such that the within-cluster sum of squared distances is minimized:
\begin{equation*}
P^* = \mathrm{arg}\min_P \sum_{j=1}^k \sum_{\vec{x}_i \in p_j} dist(\vec{x}_i, \vec{c}_j)^2,
\end{equation*}
where $\vec{c}_j$ is the centroid of partition $p_j \in P$. However, in the Euclidean space this is an NP-hard optimization problem for $k \geq 2$, even for number of dimensions $m = 2$. For this reason, heuristic methods to find the local optimum such as K-means have been proposed. 
The K-means clustering model is a popular data-mining technique used to segment data points into groups, each containing data points similar to one another and dissimilar to data points in other groups, \cite{1056489}. Similar to the original definition, it uses an iterative algorithm that continuously readjusts until the best possible segmentation is achieved. In each iteration of the algorithm, each record is assigned to the cluster whose center is closest.
In the context of time series, \cite{Aghabozorgi2015} argues that their unique characteristics make them unsuitable to conventional clustering algorithms. In particular, the high dimensionality, very high feature correlation, and typically large amount of noise have been viewed detrimental to their performance. Thus, research has focused on either (a) representing time series in a lower dimension compatible with conventional clustering algorithms or (b) use distance measures based on raw time-series or its representations. The common thread in both approaches is clustering of the transferred, extracted or raw time-series using conventional clustering algorithms such as k-means. However, \cite{Paparrizos2017-nw} identify three main drawbacks: (i) they cannot easily scale to large volumes of data, (ii) they are domain-specific or only work for specific data sets, and (iii) they are sensitive to outliers and noise. 

\subsection{Topological Data Analysis}
Topological Data Analysis (TDA) is a recent field that emerged from a combination of various statistical, computational, and topological methods during the first decade of the century. It allows to find shape-like structures in the data and has proven to be a powerful exploratory approach for noisy and multi-dimensional data sets. For a detailed introduction, the reader is invited to consult \cite{chazal2017introduction}. \cite{Turner2019-bc} highlights that TDA is usually concerned with analyzing complex data with a complicated geometric or topological structure. It is possible to represent this structure with a family of topological spaces, a filtration, defined as $\{K_a\}_{a \in A \subset \mathbb{R}}$ if $K_a \subset K_b$ whenever $a \leq b$. The inclusion of $K_a \subset K_b$ induces a homomorphism between the homology groups $H_k(K_a)$ and $H_k(K_b)$. The persistent homology is an image of $H_k(K_a)$ in $H_k(K_b)$, it encodes the $k$-cycles in $K_a$ that are independent with respect to boundaries in $K_b$. Thus, $H_k(a,b):= \frac{Z_k(K_a)}{(B_k(K_b) \cap Z_k(K_a))}$ with $Z_k$ as the cycle group and $B_k$ as the boundary group, both subgroups of the $k$-th chain group $C_k$ of $K$, a free Abelian group on its set of oriented $k$-simplices. The popular representations of persistent homology information are the barcode and the persistence diagram. A barcode is a collection of intervals [birth,death) each representing the birth, and death, values of a persistent homology class. This collection of intervals satisfies the condition that for every $a \leq b$, the number of intervals containing $[a,b)$ is $dim(H_k(a,b))$. A persistence diagram is the multi-set of points in the plane where each bar in the barcode is sent to the point with first coordinate, its birth time, and its second coordinate, its death time. After a filtration of topological spaces is built from the observations, a persistent homology is applied, a commonly used summary statistic. This filtration can be summarized in terms of the evolution of the homology. Thus, a summary from a single complex object is created. This object can be a point cloud, a graph, a time series, etc. A wide array of topological summaries can be computed directly from a persistence diagram or barcode. Each of these is a different expression of the persistent homology in the form of a topological summary statistic.
\section{Clustering Ensemble}\label{sec:cluster_ensemble}
\begin{figure}[h]
  \centering
  \includegraphics[width=\linewidth]{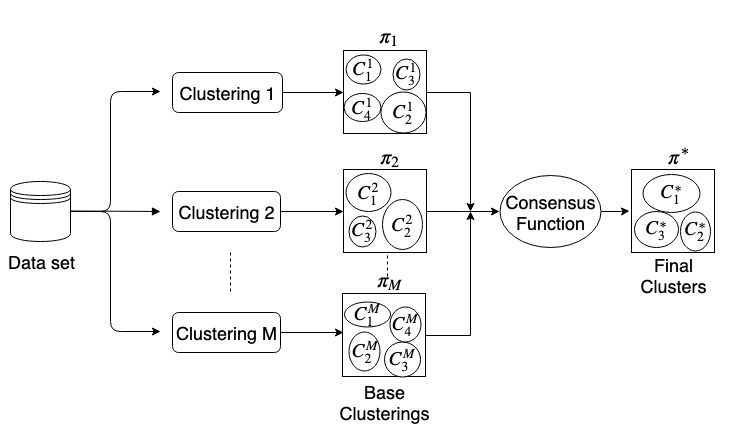}
  \caption{Process of Cluster ensemble, generate M base clusterings, then combine these solutions using the Consensus Function to get the final clusters of the data $\pi^*$ }
  \label{fig:process}
\end{figure}

The main objective of a clustering ensemble is to combine different base clusterings from a data set into a one that improves robustness and quality of clustering results, \cite{Strehl2002ClusterEA}. The intuition behind a cluster ensemble is depicted in \autoref{fig:process}. Formally, it can be described as having a data set $X=\{x_1,...,x_N\}$ which one wants to cluster into k clusters. Then, $\Pi = \{\pi_1, ..., \pi_M\}$, is a set of $M$ base clusterings that will be part of the Ensemble, where each $\pi_g$ represents a base clustering of $X$ that returns $k_g$ clusters, $\pi_g =\{C_1^g, ..., C_{k_g}^g\}$ ($k_g$ is the number of clusters in the $g$-th base clustering). For each $x_i \in X$ , $C^g(x_i)$ denotes the cluster label in the gth base clustering to which data point $x_i$ belongs. The main objective of this algorithm is to find a new final partition $\pi^*=\{C_1^*,\ldots,C_k^*\}$, where $k$ denotes the number of clusters in the final clustering result, of a data set $X$ that summarizes the information from the cluster ensemble $\Pi$.

\subsection{Voting Ensemble based on Multivariate Gaussian Mixture Models}\label{sec:gmm_voting}
This work proposes two novel methods for clustering ensemble. The first one is done through a voting ensemble based on Multivariate Gaussian Mixture Models. This technique has been named \textit{GMM\_Voting}. It seeks to find the partition $\pi^*$ by first finding the frequency in which a point is assigned to a cluster across all base clusterings and then by applying a Gaussian Mixture Model to find the final partition.
The proposed clustering ensemble differs from earlier approaches. It can not be classified under the so-called \textit{voting methods}, a category of techniques, where one firstly solves the label correspondence problem so that the labels are consistent through the base clusterings, then applies plurality voting system where each of the base clustering votes for a data point to be assigned to a specific cluster, thus the data point is assigned to the cluster where it was assigned for most of the base clustering. The method cannot also be considered part of the family of Mixture Models (MM) techniques, where they intend to solve the problem by finding the probability of assigning labels to the objects in the partitions using finite MM. The proposed approach uses the Relabeling-Voting Matrix solely as a representation of each data point and then applies a Gaussian Mixture Model (GMM); in contrast to the voting approach proposed in \cite{1410288}. After having calculated the Relabeling-Voting Matrix, they assign each data point to the cluster with the highest votes.
Compared to other MM approaches, \textit{GMM\_Voting}, differs in that other MM approaches use directly the labels from the base clustering as features for the MM without a previous processing. For example, \cite{1524981} uses the label vectors representing data points and then uses a MM to fit non-Gaussian distributions to the data. \cite{YU201416} generates base clusterings with a GMM first and then applies the Bhattacharyya distance function to calculate the distance between two components and create the representative matrix. Lastly, an ensemble graph is built. In \cite{10.1007/978-3-319-12640-1_1}, they make use of the GMM technique to calculate the probability of assigning a data point to a candidate cluster for each clustering method. After that, they use Dempster–Shafer theory, \cite{shafer1976mathematical}, to decide the final result based on the sum of confidences from different clustering methods.

\subsubsection{Relabeling-Voting}
The first step in the ensemble algorithm is to calculate the Relabel-Voting matrix. This matrix gives information about how frequently a point in the data is assigned to each cluster. In this step, the voting matrix is calculated following the method proposed by \cite{1410288}. 
First, it is necessary to have consistency in the labels of each clustering. Thus, the label $C^1_1$ from clustering $\pi_1$ should mean the same as $C^2_1$ from $\pi_2$. 
In order to achieve the most consistent labeling of clusters in every clustering, it is required to solve a correspondence problem equivalent to the maximum weight bipartite matching problem. It is necessary to select a clustering as the reference clustering $\pi_r$ from the $M$ partitions in an ensemble $\Pi$ , i.e., $\pi_r \in \Pi$.
Once the reference clustering is defined, one can create a contingency matrix $\Omega \in R^{k\times k}$ containing a number of cluster label co-occurrences counted for $\pi_r$ and other clusterings $\pi_g$ from the ensemble, where $k$ is the number of clusters in each partition. 
Each entry $\Omega (l , l')$ denotes the co-occurrence statistics between labels $l \in \pi_r$ and $l' \in \pi_g$, is defined by $\Omega (l , l') = \sum_{\forall x_i \in X} w(x_i)$
where $w(x_i)=1$, if $ C^r(x_i)=l$ and $C^g (x_i ) = l'$, otherwise $w(x_i)=0$. 
Further, $\Theta(l, l')$ must be defined. $\Theta \in R^{k \times k}$ is another matrix representing correspondences among labels of partitions $\pi_r$ and $\pi_g$. An entry $\Theta(l, l') = 1$, if label $l \in \pi_r$ corresponds to label $l' \in \pi_g$ , $0$ otherwise.
Having obtained $\Omega$ and $\Theta$ , the label correspondence is solved by maximizing their sum of sums $\sum_{l=1}^k\sum_{l'=1}^k \Omega(l,l')\Theta(l,l')$.
The solution can be found using the Hungarian algorithm, \cite{doi:10.1137/0105003}, Then, each of the $M-1$ remaining partitions is re-labeled with respect to the chosen $\pi_r$, by following the previously mentioned steps. Hence, a globally consistent label set is employed to all partitions. With this, a plurality voting can be employed to estimate the Relabel-Voting Matrix $ RV \in N^{N \times k}$.

\subsubsection{Multivariate Gaussian Mixture Model}
During the second step, the algorithm finds the probability of each data point $x_i \in X$ to be in a cluster $C_i^*$ from $\pi^*$.
It takes as a representation of each data point $x_i$ the vectors from the Relabel-Voting Matrix $RV_{x_i}$ with $RV_{x_i} = [RV^1(x_i),\ldots, RV^k(x_i)]$, where $RV^k(x_i)$ is the number of votes for the $x_i$ point in the $k$-th cluster. This can be described as a mixture of $k_{max}$ multivariate Gaussians densities where $k_{max}$ is an input parameter indicating the maximum number of Gaussian distributions that the mixture model will adjust. It is required to find $ p(C^*_k | RV_{x_i}, \hat{\phi}, \hat{\mu}, \hat{\Sigma}) $, where $\hat{\mu}$ is the component means, $\hat{\Sigma}$ the co-variances and $\hat{\phi}$ are the mixture component weights with the constraint that $\sum^{k_{max}}_{i=1}\phi_i=1$ the total probability distribution normalizes to $1$. Thus, $p(k)=\sum_{i=1}^{k_{max}} \phi_i N(x|\mu_i, \Sigma_i)$ and
$N(x|\mu_i, \Sigma_i)= \dfrac{1}{\sqrt{(2\pi)^k|\Sigma_i|}} exp \left \{- \frac{1}{2}(x-\mu_1)^T \Sigma_i^{-1}(x-\mu_i) \right \}$.
The GMM is formulated as a maximum likelihood estimation problem, which aims to find the best fitting mixture density for the given data. An Expectation-Maximization algorithm is applied to maximize the likelihood function. Finally, the final parameter for finding the conditional probability $p(C^*_k | RV_{x_i}, \hat{\phi}, \hat{\mu}, \hat{\Sigma})$ of each point in the data is used and the label corresponding to the $C^*_k$ with the highest probability is assigned. An example of this can be observed in \autoref{CE_GMM}.

\subsection{Multivariate Gaussian Mixture Model based Pairwise-Similarity}
In addition to the clustering ensemble previously presented, this work introduces a second method called \textit{GMM\_Pair}. The difference with regards to the method presented previously is that it calculates the frequency in which two data points are assigned to the same cluster across all base clusterings and then applies a GMM to find the final partition. Therefore, on a first step, the Pair-wise similarity matrix is computed as $CO(x_i, x_j) = \sum_{g=1}^M S_g(x_i,x_j)$
where, $S_g(x_i, x_j)= 1$ if $C^g(x_i)=C^g(x_j)$ and $0$ otherwise. The second step computes a GMM.
\section{Clusterwise Regression}\label{sec:clusterwise}
According to \cite{Gitman2018-lb}, in clusterwise regression, given a data set $X \in \mathbb{R}^{n \times d}, Y \in \mathbb{R}^{n}$ with $n$ points and $d$ features, one tries to find a partition of the data into $k$ disjoint clusters that minimizes the sum of squared errors of linear regression models inside each cluster. Thus,
\begin{equation*}
\begin{aligned}
& \underset{C_i, w_i, b_i}{\text{minimize}}
  & & \sum^k_{i=1} \sum_{j \in C_i} (y_j - x^T_j w_i - b_i)^2
        \\
& & &   \text{s.t.} \cup^k_{i=1} C_i = {1, \ldots, n}, C_i \cap C_j = \emptyset\, \text{for}\, i \neq j    \,.
\end{aligned}
\end{equation*}
Clusterwise regression has known limitations such as being a NP-hard problem and computationally slow. To overcome this problematic, this research is inspired by the approach proposed by \cite{Gitman2018-lb} and adapts it for time series data. The clusterwise regression is done with fixed labels and it alternates between two steps: regression and labeling. During the regression step, an individual model is fit into each cluster. On the labeling step, each observation with a training error exceeding the average of its respective cluster is re-assigned to the cluster with the best result. This is done iteratively until no further improvement can be achieved. As a next step, to predict test labels, a classifier is trained on the objects $x_i$ and the corresponding clusterwise labels. Thus, unlabeled observations are assigned to the pre-existing clusters and clusterwise regression can be then done on the enlarged clusters.
\section{Matrix Factorization}
\label{sec:matrix_factorization}
Another contribution of this work is the presentation of an approach based on matrix factorization methods (MF). They are used in a variety of applications such as recommender systems, demand forecasting, \cite{Rivera_2018}, signal processing, \cite{Weng2012}, computer vision, \cite{Chen2004}, and others.

Let $Y$ be $T\times n$ sparse or dense matrix of observations of $n$ objects spanning the period of $T$ time steps, i.e. each column $i=1,\,\ldots,\,n$ of $Y$ is a times series $y^{(i)} = (Y_{ti})_{t=1}^T$ related to the $i$-th object. For instance, $Y$ may represent consumption expenditures within a longitudinal study of households, the hourly records of electricity consumption at different substations, financial time series or changes in stock levels. The problem of factorizing a fully or partially observed $T\times n$ matrix $Y$ consists of finding $d$-dimensional factors $Z$ and the corresponding factor loadings $F$, in the form of $T \times d$ and $d \times n$ matrices respectively, such that their product $Z F$ most accurately recovers the observed $Y$, i.e. $Y_{ti} \approx \sum_{j=1}^d Z_{tj} F_{ji}$. This is usually achieved by solving the following optimization problem:
\begin{equation} \label{eq:general_mf}
\begin{aligned}
& \underset{F, Z}{\text{minimize}}
  & & \tfrac1{2 \lvert \Omega\rvert}
        \|\mathcal{P}_\Omega(Y - Z F)\|^2 
        \\
& & &   + \lambda_F \mathcal{R}_F(F) 
        + \lambda_Z \mathcal{R}_Z(Z)
      \,,
\end{aligned}
\end{equation}
where $\Omega\subset \{1..T\} \times \{1..n\}$ is the sparsity pattern of $Y$, $\mathcal{P}_\Omega$ zeroes out unobserved entries. The coefficients $\lambda_F$ and $\lambda_Z$ are non-negative regularization coefficients which govern the trade-off between the reconstruction error and the regularizing terms $\mathcal{R}_F$ and $\mathcal{R}_Z$. The latter depends on the particular desired properties of the factorization, such as sparsity or row-wise group sparsity, \cite{nazarovetal2018}, typically in conjunction with a Ridge regression-type penalty ($\ell^2$ norm). The key issue with~\eqref{eq:general_mf} is that without extra structural requirements on $Z$ it is impossible to apply this technique to time series prediction. A recent paper, \cite{yuetal2016}, proposes a novel regularization term $\mathcal{R}_Z$ for~\eqref{eq:general_mf}, that enables forecasting beyond $T$ by imposing autoregressive time-series properties on the latent factors $Z$. The corresponding optimization problem, which imposes $AR(p)$ (autoregression of order $p$) dynamics on the factors, is
\begin{small}
\begin{equation} \label{eq:trmf}
\begin{aligned}
& \underset{F, Z, \phi}{\text{minimize}}
  & & \tfrac1{2 \lvert \Omega\rvert}
        \|\mathcal{P}_\Omega(Y - Z F)\|^2
      + \tfrac{\lambda_F}2 \tfrac1{d n} \|F\|^2
      + \tfrac{\lambda_\phi}2 \tfrac1{d p} \|\phi \|^2
      \\
& & & + \tfrac{\lambda_Z}2 \Bigl(
            (1 - \eta_Z) \tfrac1{T d} \|Z\|^2
    \\
& & & + \eta_Z \tfrac1{(T - p) d}
            \sum_{j=1}^d \sum_{t=p+1}^T \bigl(
                Z_{tj} - \sum_{i=1}^p \phi_{ji} Z_{t-i,j}
            \bigr)^2
        \Bigr)
      \,.
\end{aligned}
\end{equation}
\end{small}
In this formulation $\lambda_\phi$ regularizes and stabilizes the estimates of autoregression coefficients $\phi$, that determine the dynamics of the latent factor $Z$, which are $d$-dimensional time series. The parameter $\eta_Z\in [0, 1]$ regulates the relative contribution of the ridge-like and forecastability penalties to the estimation of the factor series $Z$.
Forecasting beyond the last observation in $Y$ is done using the estimated factors $Z$ and the parameters $\phi$ of their autoregressive dynamics. If $z^{(j)} = (Z_{tj})_{t=1}^T$ is the time series of the $j$-th latent factor, then its $h$-step ahead dynamic forecast beyond time $T$ is calculated using
\begin{equation} \label{eq:factor_dyn_forecast}
\hat{z}^{(j)}_{T+h\mid T}
  = \phi_{j1} \hat{z}^{(j)}_{T+h-1\mid T}
    + \cdots + \phi_{jp} \hat{z}^{(j)}_{T+h-p\mid T}
  \,,
\end{equation}
with $\hat{z}^{(j)}_{T+h-k\mid T} = z^{(j)}_{T+h-k}$ for any $k \geq h$. Based on~\eqref{eq:factor_dyn_forecast} and the form of the matrix factorization, the $h$-step ahead forecast for the $i$-th object's time series is given by $\hat{y}^{(i)}_{T+h\mid T} = \sum_{j=1}^d \hat{z}^{(j)}_{T+h\mid T} F_{ji}$.
According to~\cite{yuetal2016}, the problem~\eqref{eq:trmf} could be further modified to incorporate hierarchical relations among the columns of $Y$ by adding a structured $\ell^2$-like regularization on the loadings $F$. If $A$ is the matrix of object similarities, that correspond to the columns of $Y$, then the regularizer on $F$ could be
\begin{small}
\begin{equation} \label{eq:trmf_graph_loadings}
  \tfrac{\lambda_F}2 \Bigl(
    (1 - \eta_F) \tfrac1{d n} \|F\|^2
    + \eta_F \tfrac1{d n} \bigl\| F (I - A^{\mathrm{T}} D^{-1}) \bigr\|^2
  \Bigr)
  \,,
\end{equation}
\end{small} where $D$ is the degree matrix of the graph corresponding to $A$, and the parameter $\eta_F \in [0, 1]$ balances the ridge-like penalty and hierarchical regularization, which encourages columns in $F$ to be close to each other, if the corresponding objects are similar according to $A$.
\section{Topological RFM}
\label{sec:topological_rfm}
\begin{figure*}[!htb]
\includegraphics[width=\columnwidth*2]{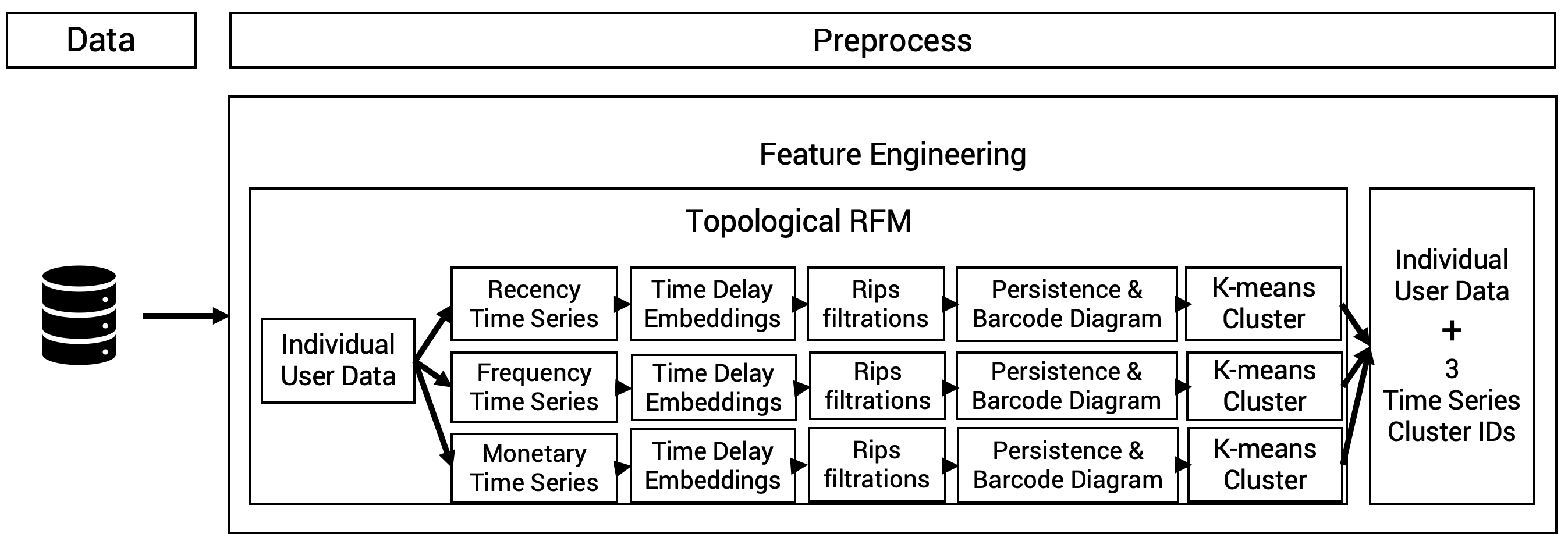}
\caption{\textit{Topological RFM} consisting of 3 time series per user and 3 instances of Time Delay Embeddings, Rips Filtrations, Persistence diagrams and Barcode diagrams and K-means clusters}
\label{fig:methods:process_diagram_4}
\end{figure*}

This work discusses an extension to \textit{Topological RFM}, \cite{Rivera-Castro2019-qk}. 
The method generates three time series from a seed time series and replicates the RFM framework. Thus, it creates a time series for Recency, last time a user event took place, Frequency, how often a user has triggered an event, and Monetary, the financial value of the events generated by the user.
\autoref{fig:methods:process_diagram_4} shows the architecture proposed for \textit{Topological RFM} and \autoref{repro:topo_rfm} details its respective five steps. This structure is necessary to convert time series data into an object that can be used by a topological data analysis algorithm.
\textit{Topological RFM} is enhanced by converting it into a clusterwise regression method. To achieve this, two cluster ensemble methods were introduced in \autoref{sec:cluster_ensemble}. They are used to combine the three clusters generated by RFM, whereas in the original work, a heuristic was used. This research adds further novelty by using a matrix factorization method for demand forecasting, discussed in \autoref{sec:matrix_factorization}, as the regressor for the proposed clusterwise regression. To the best knowledge of this work, matrix factorization methods have not been used previously within the clusterwise regression context nor cluster ensembles for TDA clustering of time series.

\section{Data Sets}\label{sec:dataset}
\subsection{CDNow}\label{sec:cdnow}
\begin{figure}[!htb]
\begin{center}
\includegraphics[width=1\columnwidth]{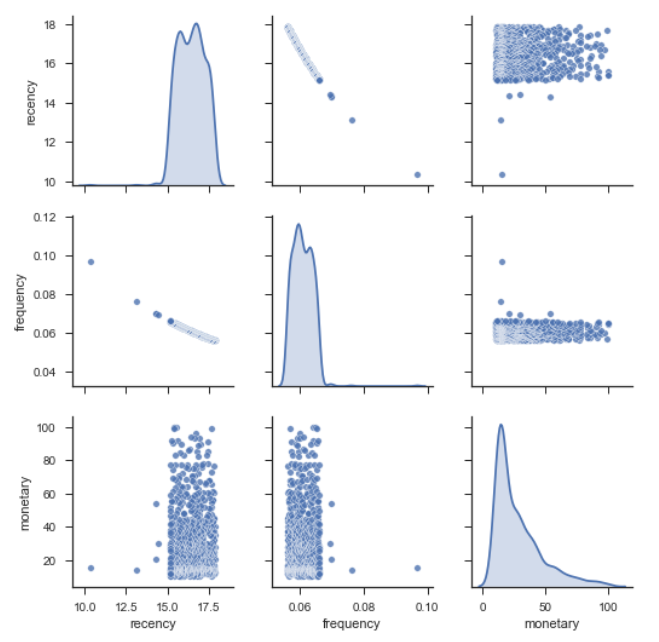}
\end{center}
\caption{Distribution of the Recency Frequency Monetary values for the CDNow data set.}
\label{fig:rfm:cdnow}
\end{figure}

The CDNow data set has been commonly used in the CRM literature. It contains the entire purchase history of a cohort comprising 23,570 individuals from their first purchase in the first quarter of 1997 up to the end of June 1998. This is well reflected on the recency plot in \autoref{fig:rfm:cdnow}, almost everyone in the cohort did not make a purchase recently and this has been their only purchase according to the frequency plot. Similarly, the monetary value is not large. This coincides well with the type of product sold at the store, compact discs with music. The data set can be downloaded \footnote{\url{http://www.brucehardie.com/datasets/}}.

\subsection{Cloud Computing Provider}\label{sec:cloudcomputing}
\begin{figure}[!htb]
\begin{center}
\includegraphics[width=1\columnwidth]{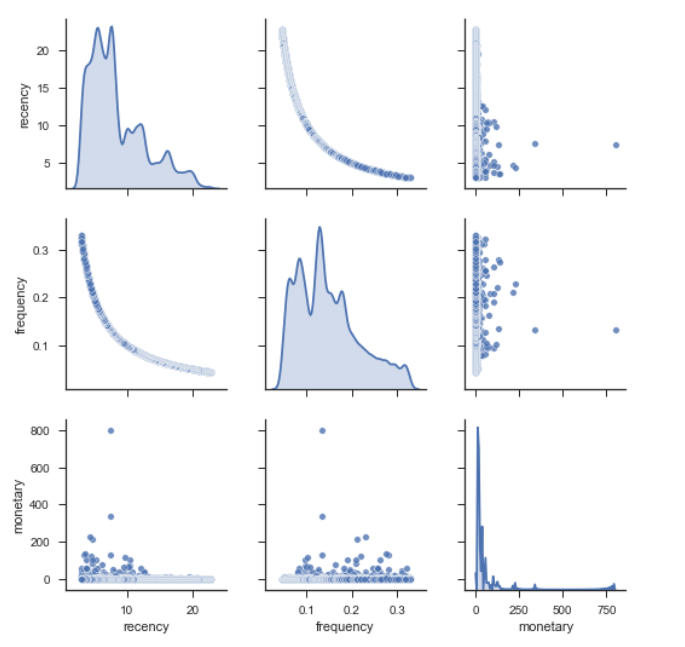}
\end{center}
\caption{Distribution of the Recency Frequency Monetary values for the Cloud Computing data set.}
\label{fig:rfm:cloud}
\end{figure}

\begin{figure}[!htb]
\begin{center}
    \includegraphics[width=1\columnwidth]{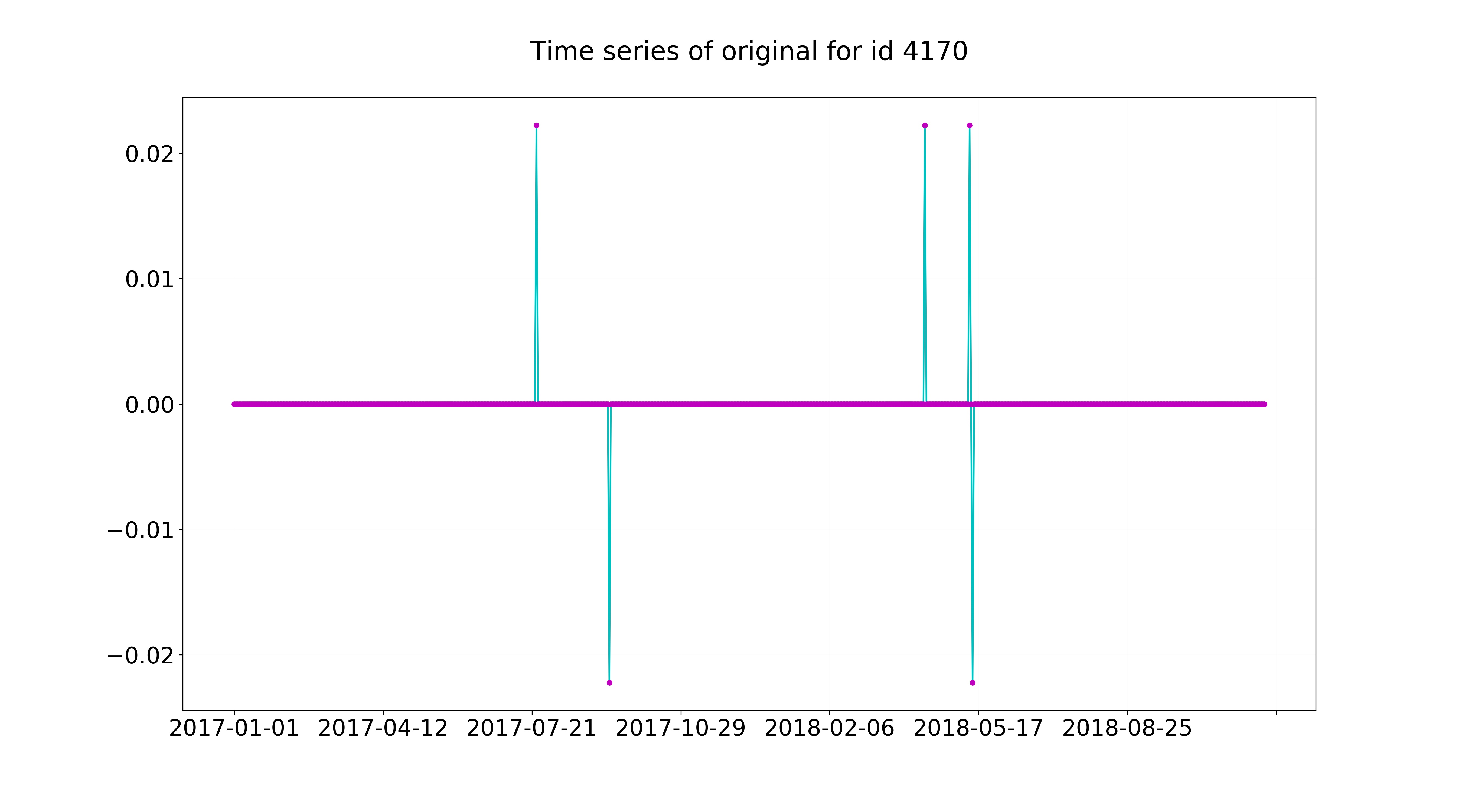}
\end{center}
\caption{Time Series of user 4170 from the Cloud Computing data set}
\label{fig:ts_rfm:ts4170}
\end{figure}

This study presents a customer segmentation method developed for a cloud computing provider. The data represents a small subset of the customer base. It contains observations with time stamps documenting whenever a customer has booked computing (CPU) time with the provider between 2017 and 2018, the duration and the type of product booked. The nature of the cloud computing industry is reflected in the RFM distribution seen in \autoref{fig:rfm:cloud}. Most customers have used the service recently either rarely or frequently. This can be explained by two common use cases in cloud computing. The first one consists of booking a computing instance and leave it running i.e., to service an Internet website. The second use case is to book cloud computing on demand for a specific task and once it is finished, shut down the instance. This is also seen in the monetary plot, where again two use cases appear. One corresponds to the occasional user with a small budget and the other one to the intensive user of computing services.
As a further example, \autoref{fig:ts_rfm:ts4170} shows the demand of a randomly selected user. As discussed previously, in this data set, there is a strong presence of sporadic users with few values.

\section{Experiments} \label{sec:experiments}
To validate the proposed methods in this work, this study carried out an assessment consisting of the following setting described previously in \autoref{fig:general:process_diagram_1}:
\begin{enumerate}
\item Fit Matrix Factorization on the full data set
\item Fit Baseline on the full data set,
\item Fit clusterwise regression with Matrix Factorization using \textit{GMM\_Vote} clusters,
\item Fit clusterwise regression with Matrix Factorization using \textit{GMM\_Pair} clusters,
\item Fit clusterwise regression with Theta Method using \textit{GMM\_Vote} clusters,
\item Fit clusterwise regression with Theta Method using \textit{GMM\_Pair} clusters.
\end{enumerate}

Each of these experiments were done in batches of 2600, 5000 and 10000 time series.
In the cases when the model was fit on the full data set, the data was divided into a training and test set using a 70\%/30\% split.
In the case of the clusterwise regression, a different approach was followed according to the description in \autoref{sec:clusterwise}; this looks as following.
(a) 2000 time series were taken to train the clusterwise regression and its classifier. 
(b) Out of those, 1400 (70\%) were used for the clusterwise regression.
(c) The generated cluster labels were then used to train the classifier, a gradient boosting tree. 
(d) To test it, the other 600 (30\%) time series were used.
(f) Whereas the amount of time series for the clusterwise regression and its classifier remained constant throughout the experiments, the number of unlabeled time series used to evaluate the methods was increased in batches of 600, 3000 and 7000 respectively.
(g) The testing phase consisted on taking each batch and predicting their labels using the classifier. With the labels, the time series were assigned to their respective clusters and the clusterwise regression was used to predict them.
(h) As the data in question are time series, both predictive models, were trained with 70\% of the observations of a given time series and tested with the other 30\%.
To compare the quality of the results, Root Mean Square Error (RMSE) was used. This is defined as $RMSE = \small{ \sqrt{\frac{1}{T}\sum_{t=1}^T (x_{1,t} - x_{2,t})^2}}$. The results of the experiment can be found in \autoref{tab:models:1}.

\subsection{Baseline}\label{sec:baseline}
As a baseline for comparisons, the \textit{Theta method} is used. It decomposes a time series into two curves and aims at magnifying the short- and long-term movement of the data. Empirically, it has been successful in various time series forecasting competitions. For a detailed explanation, the reader can consult, \cite{Assimakopoulos2000-sy}.

\begin{small}
  \begin{table}
\caption{Overview of results using mean RMSE. Low values are better. Best results for each batch in bold. MF: Matrix Factorization. Theta: Theta Method}
\begin{center}\label{tab:models:1}
\begin{tabulary}{\linewidth}{CCCCC}
    \toprule
Dataset	&	Model	&	Method	&	Batch	&	RMSE	\\	\hline
Cloud	&	MF	&	All data	&	2600	&	1.38	\\	
Cloud	&	Theta	&	All data	&	2600	&	1.28	\\	
Cloud	&	MF	&	GMM\_Vote	&	2600	&	1.20	\\	
Cloud	&	Theta	&	GMM\_Vote	&	2600	&	1.24	\\	
\textbf{Cloud}	&	\textbf{MF}	&	\textbf{GMM\_Pair}	&	\textbf{2600}	&	\textbf{0.43}	\\	
Cloud	&	Theta	&	GMM\_Pair	&	2600	&	0.93	\\	
Cloud	&	MF	&	All data	&	5000	&	1.49	\\	
Cloud	&	Theta	&	All data	&	5000	&	2.44	\\	
\textbf{Cloud}	&	\textbf{MF}	&	\textbf{GMM\_Vote}	&	\textbf{5000}	&	\textbf{0.22}	\\	
Cloud	&	Theta	&	GMM\_Vote	&	5000	&	1.28	\\	
Cloud	&	MF	&	GMM\_Pair	&	5000	&	1.08	\\	
Cloud	&	Theta	&	GMM\_Pair	&	5000	&	1.12	\\	
Cloud	&	MF	&	All data	&	10000	&	1.78	\\	
Cloud	&	Theta	&	All data	&	10000	&	1.60	\\	
\textbf{Cloud}	&	\textbf{MF}	&	\textbf{GMM\_Vote}	&	\textbf{10000}	&	\textbf{0.25}	\\	
Cloud	&	Theta	&	GMM\_Vote	&	10000	&	1.46	\\	
Cloud	&	MF	&	GMM\_Pair	&	10000	&	0.37	\\	
Cloud	&	Theta	&	GMM\_Pair	&	10000	&	1.4	\\	
CDNow	&	MF	&	All data	&	2600	&	2.72	\\	
CDNow	&	Theta	&	All data	&	2600	&	2.78	\\	
CDNow	&	MF	&	GMM\_Vote	&	2600	&	2.11	\\	
\textbf{CDNow}	&	\textbf{Theta}	&	\textbf{GMM\_Vote}	&	\textbf{2600}	&	\textbf{2.10}	\\	
CDNow	&	MF	&	GMM\_Pair	&	2600	&	2.40	\\	
CDNow	&	Theta	&	GMM\_Pair	&	2600	&	2.41	\\	
CDNow	&	MF	&	All data	&	5000	&	2.59	\\	
CDNow	&	Theta	&	All data	&	5000	&	2.63	\\	
\textbf{CDNow}	&	\textbf{MF}	&	\textbf{GMM\_Vote}	&	\textbf{5000}	&	\textbf{1.32}	\\	
CDNow	&	Theta	&	GMM\_Vote	&	5000	&	2.59	\\	
CDNow	&	MF	&	GMM\_Pair	&	5000	&	2.40	\\	
CDNow	&	Theta	&	GMM\_Pair	&	5000	&	2.41	\\	
CDNow	&	MF	&	All data	&	10000	&	3.14	\\	
CDNow	&	Theta	&	All data	&	10000	&	2.78	\\	
\textbf{CDNow}	&	\textbf{MF}	&	\textbf{GMM\_Vote}	&	\textbf{10000}	&	\textbf{1.28}	\\	
CDNow	&	Theta	&	GMM\_Vote	&	10000	&	2.35	\\	
CDNow	&	MF	&	GMM\_Pair	&	10000	&	2.26	\\	
CDNow	&	Theta	&	GMM\_Pair	&	10000	&	2.27	\\	
 \bottomrule
 \end{tabulary}
 \end{center}
 \end{table}
 \end{small}
 \section{Discussion and Learnings}\label{sec:discussion}
This work seeks to expose the practitioner to state of the art methods in the areas of Topological Data Analysis and Time Series Clustering applied to problems in Customer Relationship Management and demand planning by proposing a machine learning system that can carry out both tasks. 
\cite{Rivera-Castro2019-qk} show that the RFM framework reproduced using TDA and time series, \textit{Topological RFM}, offers a superior performance over similar alternatives and over its original inspiration, the Recency Frequency Framework. This work takes this model as a base with the objective of providing a unique tool for customer segmentation and demand forecasting that can be used across business departments. To achieve this, it proposes to combine the RFM clusters using the clustering ensembles presented in \autoref{sec:cluster_ensemble} and fit these results into the clusterwise regression methodology described in \autoref{sec:clusterwise} and the matrix factorization technique in \autoref{sec:matrix_factorization}. Given that other methods for time series prediction can be used in the clusterwise process, as a benchmark the \textit{Theta method} described in \autoref{sec:baseline} was used. 
Section \ref{sec:experiments} carried out a set of experiments to assess the suitability of this system. The results show that a clusterwise regression with matrix factorization consistently outperformed the baseline. This is specially the case once the data sets started to be increased. For example, for the cloud computing data set a RMSE of 0.43 was obtained with 2600 time series. This error was reduced to 0.22 once the batch size increased to 10000 time series. Similarly, for the CDNow data set, the error was reduced from 2.10 for the Theta method and 2.11 for the MF method with 2600 time series to 1.28 using 10000 time series. It is worth noting that the performance of the clustering ensemble can vary greatly. For this set of experiments, \textit{GMM\_Vote} gave largely the best results. However, in some cases, \textit{GMM\_Pair} can also be a viable option. 
As it is the case in clusterwise regression, improving the results came at the cost of long computing times. The method is significantly slower than fitting the data set directly. It is to be expected that the performance will only drop as the data set increases its size. 
Similarly, Topological RFM is also not suited for large data sets. Computing homologies is expensive. Another limitation of this method is the lack of established best practices for defining the size of the sliding window to generate the point cloud. This can be especially difficult whenever dealing with sparse time series data. Thus, in order to improve the method, two bottle necks need to be addressed. On one side, the computation of the time series clusterings using TDA. On the other side, the labeling step in the clusterwise regression.

\section{Takeaways for the practitioner}
This work has as its main audience the practitioner in quantitative marketing or demand planning. Thus, it presented a full system suited to segment and score her customer base and forecast demand for each individual customer. This is attractive as both CRM and demand planning problems remain perennial in the industry. However, both tasks are usually carried out by separate departments and individuals with different profiles. This work provides a viable option to address both of them at the same time.
For the practitioner to make best use of this method, she should first assess if her time series are suited for TDA clustering. Extremely sparse data is not adequate for this technique. 
Further, it is well-known that the initialization of a clustering method can provide vastly different outcomes each time. \cite{Gitman2018-lb} suggest to run multiple initializations and average their results. However, the clustering ensembles presented here can also be used with the same objective. The practitioner can have variations of Topological RFM with changes such as the size of the sliding window or the method to do the cluster itself and then proceed to obtain a \textit{super cluster} using a clustering ensemble such as \textit{GMM\_Vote}. 
Similarly, the practitioner has to pay attention to the balance between the number of time series used to train the clusterwise regression, the classifier and the final amount used to predict. The combination of clusterwise regression plus classifier allows the method to handle medium-sized and even large data sets. It is possible to observe a drop in performance once the amount of time series to predict far exceeds the original training set.
A final benefit for the practitioner is the data set provided by this study. It is unique and novel. There are no other public data sets for B2B customer relationship management or demand planning in the cloud computing sector.
\section{Further directions}
As a next step, this work will seek to cover very large data sets. Both TDA clustering and clusterwise regression are computationally intensive and only suited for medium-sized data sets. An option is to use ideas of \cite{Lacombe2018-od} and combine TDA with Optimal Transport to speed up the computation of persistence diagrams. 
Other possibility is to consider using Optimal Transport methods for the labeling process during the clusterwise regression. 
Another line of research is to evaluate various initialization methods to define guidelines and best practices for time series clustering with TDA and clusterwise regression. A clear example of this is establishing the best sizes for sliding windows under varying conditions. 
Overall, TDA is a nascent field and to the best of the knowledge of this study, on the algorithmic side, this is the first work combining TDA for time series clustering with clustering ensembles, clusterwise regression and matrix factorization techniques. On the application side, this novel combination is shown as an effective combination to carry out CRM and demand forecasting together; thus, bridging the gap between the marketing and demand planning functions within a company. As the field grows in popularity and new business applications appear, it is to be expected that TDA clustering combined with clusterwise regression will become an essential tool for the practitioner.

\ifCLASSOPTIONcompsoc
  \section*{Acknowledgments}
\else
  \section*{Acknowledgment}
\fi
The research in \autoref{sec:literature_review}, \autoref{sec:cluster_ensemble} and \autoref{sec:matrix_factorization} was supported by the Ministry of Education and Science of the Russian Federation (Grant no. 14.756.31.0001). Other sections were supported by the Mexican National Council for Science and Technology (CONACYT), 2018-000009-01EXTF-00154.


\bibliographystyle{IEEEtran}
\bibliography{bib}

\pagebreak
\newpage

\section{Reproducibility}
\subsection{Cluster Ensembles}\label{repro:cluster_ensemble}
This work presented two novel cluster ensembles, \textit{GMM\_Voting} and \textit{GMM\_Pair}. The respective algorithms are defined as following:

\begin{algorithm}[H]
\caption{\textit{GMM\_Voting}}\label{GMM_Voting}
\hspace*{\algorithmicindent} \textbf{Input:} $\Pi$, $k_{max}$(max number of clusters)\\
\hspace*{\algorithmicindent} \textbf{Output:} $\pi^*$
\begin{algorithmic}
\\ %
\textbf{begin} \\
\textbf{Relabeling-Voting}\\
     \quad Select $\pi_r \in \Pi$ \\
     \quad Maximize $\sum_{l=1}^k\sum_{l'=1}^k \Omega(l,l')\Theta(l,l')$\\
     \quad Compute RV Matrix\\
\textbf{Multivariate GMM}\\
 \quad \textbf{Initialize} $\hat{\phi}$, $\hat{\Sigma}$, $\hat{\mu}$\\
\quad \textbf{do until} $\eta<dif$: \text{($\eta$ is a convergence criterion)} \\
    \quad \quad \textbf{Expectation}\\
    \quad \quad \quad \text{Compute} $p(C^*_k | x^{RV}_i, \hat{\phi}, \hat{\mu}, \hat{\Sigma})$ \\
    \quad \quad \textbf{Maximization}\\
    \quad \quad \quad \text{Update } $\hat{\phi}$, $\hat{\Sigma}$, $\hat{\mu}$ \\
    \quad \quad \quad \text{Compute $loglikelihood$}\\
    \quad \quad $dif=loglikelihood - loglikelihood\_prev$ \\
\quad \textbf{end}\\
$\pi^*(x_i)=\epsilon$, where $\epsilon \in [0, \ldots, k_{max}]$ is the index of component with the largest expected value, 
$i=1,\ldots,N$ \\
\textbf{return} $\pi^*$ \\
\textbf{end}
\end{algorithmic}
\end{algorithm}

\begin{algorithm}[H]
\caption{GMM\_Pair}\label{GMM_Pair}
\hspace*{\algorithmicindent} \textbf{Input:} $\Pi$, $k_{max}$ (max number of clusters)\\
\hspace*{\algorithmicindent} \textbf{Output:} $\pi^*$
\begin{algorithmic}[1]
\\ %
\textbf{begin} \\
\textbf{Pairwise-Similarity}\\
\quad Compute $CO(x_i, x_j)$\\
\textbf{Multivariate GMM}\\
\quad \textbf{Initialize} $\hat{\phi}$, $\hat{\Sigma}$, $\hat{mu}$\\
\quad \textbf{do until} $\eta<dif$: \text{($\eta$ is a convergence criterion)} \\
   \quad  \quad \textbf{Expectation}\\
   \quad \quad  \quad \text{Compute} $p(C^*_k | x^{CO}_i, \hat{\phi}, \hat{\mu}, \hat{\Sigma})$ \\
    \quad \quad \textbf{Maximization}\\
   \quad \quad  \quad \text{Update } $\hat{\phi}$, $\hat{\Sigma}$, $\hat{mu}$ \\
  \quad  \quad  \quad \text{Compute $loglikelihood$}\\
   \quad  \quad $dif=loglikelihood - loglikelihood\_prev$ \\
\quad \textbf{end}\\
$\pi^*(x_i)=\epsilon$, where $\epsilon$ is the index of component with largest expected value, 
$i=1,...,N$ \\
\textbf{return} $\pi^*$ \\
\textbf{end}
\end{algorithmic}
\end{algorithm}

\begin{table}[H]
\caption{Example of Clustering Ensemble using \textit{GMM\_Voting}}
\centering
\label{CE_GMM}
\begin{tabular}{|c|c|c|c|c|c|}
\hline
   & $\pi_1$ & $\pi_2$  & $p(C^*_1|RV_{x_i})$  & $p(C^*_2|RV_{x_i})$ & $\pi^*$\\ \hline
$x_1$ & $C^1_1$ & $C^2_1$  &   0.999   &   0.001   &  $C^*_1$ \\ \hline
$x_2$ & $C^1_1$ & $C^2_2$  &    0.943   &   0.057   &  $C^*_1$ \\ \hline
$x_3$ & $C^1_2$ & $C^2_1$  &    0.260   &   0.740   &  $C^*_2$ \\ \hline
$x_4$ & $C^1_2$ & $C^2_2$   &    0.115   &   0.885   &  $C^*_2$ \\ \hline
$x_5$ & $C^1_3$ & $C^2_2$   &    0.019   &   0.981   &  $C^*_2$ \\ \hline
\end{tabular}

\end{table}

\subsection{Topological RFM}\label{repro:topo_rfm}
The 'Topological RFM' pipeline consists of the following steps:
\begin{enumerate}
\item As a first step, three time series are generated for Recency, Frequency and Monetary respectively.
\item The time series have to be sliced. This is done using sliding windows. The objective is to generate delay embeddings that can be projected as a point cloud.
\item Once the three point clouds have been obtained, Rips filtration, a popular algorithm in TDA, is used, with the objective of obtaining death and birth complexes. These processes can be visualized in the form of persistence diagrams. The points of interest are those outside of the diagonal.
\item As a fourth step, barcode diagrams are generated for both 0- and 1- dimensional homologies. They help to visualize the birth-death filtered complexes. The focus is on the 1-dimensional homologies (loops). The 0-dimensional ones do not provide relevant information.
\item As a final step, a clustering is done using K-means based on features extracted from the barcodes. The number of clusters for each of Recency, Frequency and Monetary is decided using the Elbow method, a popular technique in cluster analysis.
\end{enumerate}

\end{document}

%% file: 00_head.tex
\ifCLASSOPTIONcompsoc
  \usepackage[nocompress]{cite}
\else
  \usepackage{cite}
\fi
\usepackage{hyperref}

\ifCLASSINFOpdf
\else
\fi
\usepackage{amsmath}
\ifCLASSOPTIONcompsoc
  \usepackage[caption=false,font=footnotesize,labelfont=sf,textfont=sf]{subfig}
\else
  \usepackage[caption=false,font=footnotesize]{subfig}
\fi

\usepackage{booktabs, tabulary}       
\usepackage{amsfonts}       
\usepackage{nicefrac}       
\usepackage{microtype}      
\usepackage{lipsum}
\usepackage{algorithm}
\usepackage{mathtools}
\usepackage[noend]{algpseudocode}
\usepackage{float}

%% file: dsaa19.bbl
\begin{thebibliography}{10}
\providecommand{\url}[1]{#1}
\csname url@samestyle\endcsname
\providecommand{\newblock}{\relax}
\providecommand{\bibinfo}[2]{#2}
\providecommand{\BIBentrySTDinterwordspacing}{\spaceskip=0pt\relax}
\providecommand{\BIBentryALTinterwordstretchfactor}{4}
\providecommand{\BIBentryALTinterwordspacing}{\spaceskip=\fontdimen2\font plus
\BIBentryALTinterwordstretchfactor\fontdimen3\font minus
  \fontdimen4\font\relax}
\providecommand{\BIBforeignlanguage}[2]{{%
\expandafter\ifx\csname l@#1\endcsname\relax
\typeout{** WARNING: IEEEtran.bst: No hyphenation pattern has been}%
\typeout{** loaded for the language `#1'. Using the pattern for}%
\typeout{** the default language instead.}%
\else
\language=\csname l@#1\endcsname
\fi
#2}}
\providecommand{\BIBdecl}{\relax}
\BIBdecl

\bibitem{Rivera-Castro2019-qk}
R.~Rivera-Castro, P.~Pilyugina, A.~Pletnev, I.~Maksimov, W.~Wyz, and
  E.~Burnaev, ``Topological data analysis of time series data for {B2B}
  customer relationship management,'' 2019, in press.

\bibitem{Zhang2015-wa}
Y.~Zhang, E.~T. Bradlow, and D.~S. Small, ``Predicting customer value using
  clumpiness: From {RFM} to {RFMC},'' \emph{Marketing Science}, vol.~34, no.~2,
  pp. 195--208, Mar. 2015.

\bibitem{Platzer2016-si}
M.~Platzer and T.~Reutterer, ``Ticking away the moments: Timing regularity
  helps to better predict customer activity,'' \emph{Marketing Science},
  vol.~35, no.~5, pp. 779--799, Sep. 2016.

\bibitem{2017arXiv170905548R}
R.~{Rivera} and E.~{Burnaev}, ``{Forecasting of commercial sales with large
  scale Gaussian Processes},'' \emph{ArXiv e-prints}, Sep. 2017.

\bibitem{Rivera_2018}
R.~Rivera, I.~Nazarov, and E.~Burnaev, ``Towards forecast techniques for
  business analysts of large commercial data sets using matrix factorization
  methods,'' \emph{Journal of Physics: Conference Series}, vol. 1117, p.
  012010, nov 2018.

\bibitem{Rivera-Castro2019-dj}
R.~Rivera-Castro, I.~Nazarov, Y.~Xiang, A.~Pletneev, I.~Maksimov, and
  E.~Burnaev, ``Demand forecasting techniques for build-to-order lean
  manufacturing supply chains,'' 2019, in press.

\bibitem{Fleisch2003}
E.~Fleisch and C.~Tellkamp, ``{Inventory inaccuracy and supply chain
  performance: a simulation study of a retail supply chain},''
  \emph{International Journal of Production Economics}, vol.~95, no.~3, pp.
  373--385, mar 2005.

\bibitem{2018esade}
N.~{Agell} and M.~{Carricano}, ``{Adopcion e impacto del Big Data y Advanced
  Analytics en España},'' \emph{ESADE Business and Law School}, May 2018.

\bibitem{2017apec}
C.~{Pompa} and T.~{Burke}, ``{Data Science and Analytics Skills Shortage:
  Equipping the APEC Workforce with the Competencies Demanded by Employers},''
  \emph{APEC Human Resource Development Working Group}, 2017.

\bibitem{Zaki2016-oj}
M.~Zaki, D.~Kandeil, A.~Neely, and J.~R. McColl-Kennedy, ``The fallacy of the
  net promoter score: Customer loyalty predictive model,'' 2016.

\bibitem{Wubben2008-yf}
M.~W{\"u}bben, \emph{\BIBforeignlanguage{en}{Analytical {CRM}: Developing and
  Maintaining Profitable Customer Relationships in {Non-Contractual}
  Settings}}.\hskip 1em plus 0.5em minus 0.4em\relax Gabler Verlag, Oct. 2008.

\bibitem{doi:10.1108/JOSM-01-2013-0018}
L.~Aksoy, ``How do you measure what you can't define?: The current state of
  loyalty measurement and management,'' \emph{Journal of Service Management},
  vol.~24, no.~4, pp. 356--381, 2013.

\bibitem{hosseini2010}
M.~Seyed~hosseini, A.~Maleki, and M.~Gholamian, ``Cluster analysis using data
  mining approach to develop crm methodology to assess the customer loyalty,''
  \emph{Expert Syst. Appl.}, vol.~37, 07 2010.

\bibitem{tirunillai2014}
S.~Tirunillai and G.~J.~Tellis, ``Mining marketing meaning from online chatter:
  Strategic brand analysis of big data using latent dirichlet allocation,''
  \emph{Journal of Marketing Research}, vol.~51, pp. 463--479, 08 2014.

\bibitem{doi:10.1108/08876041211223951}
C.~Baumann, G.~Elliott, and S.~Burton, ``Modeling customer satisfaction and
  loyalty: survey data versus data mining,'' \emph{Journal of Services
  Marketing}, vol.~26, no.~3, pp. 148--157, 2012.

\bibitem{Wubben2008-mc}
M.~W{\"u}bben and F.~Wangenheim, ``Instant customer base analysis: Managerial
  heuristics often ``get it right'','' \emph{J. Mark.}, 2008.

\bibitem{Tamaddoni2016-eo}
A.~Tamaddoni, S.~Stakhovych, and M.~Ewing, ``Comparing churn prediction
  techniques and assessing their performance: A contingent perspective,''
  \emph{J. Serv. Res.}, vol.~19, no.~2, pp. 123--141, May 2016.

\bibitem{Martinez2018-us}
A.~Mart{\'\i}nez, C.~Schmuck, S.~Pereverzyev, C.~Pirker, and M.~Haltmeier, ``A
  machine learning framework for customer purchase prediction in the
  non-contractual setting,'' \emph{Eur. J. Oper. Res.}, May 2018.

\bibitem{doi:10.1177/1094670506293810}
S.~Gupta, D.~Hanssens, B.~Hardie, W.~Kahn, V.~Kumar, N.~Lin, N.~Ravishanker,
  and S.~Sriram, ``Modeling customer lifetime value,'' \emph{Journal of Service
  Research}, vol.~9, no.~2, pp. 139--155, 2006.

\bibitem{Blattberg2008-jv}
R.~Blattberg, R.~C, {Byung-Do}, {Kim}, S.~Neslin, and N.~A, \emph{Database
  Marketing: Analyzing and Managing Customers}, Jan. 2008.

\bibitem{Ballings2012-ey}
M.~Ballings, D.~Van~den Poel, and E.~Verhagen, ``Improving customer churn
  prediction by data augmentation using pictorial {Stimulus-Choice} data,'' in
  \emph{Management Intelligent Systems}.\hskip 1em plus 0.5em minus 0.4em\relax
  Springer, 2012.

\bibitem{Coussement2008-hw}
K.~Coussement and D.~Van~den Poel, ``Churn prediction in subscription services:
  An application of support vector machines while comparing two
  parameter-selection techniques,'' \emph{Expert Syst. Appl.}, vol.~34, no.~1,
  Jan. 2008.

\bibitem{1056489}
S.~{Lloyd}, ``Least squares quantization in pcm,'' \emph{IEEE Transactions on
  Information Theory}, vol.~28, no.~2, pp. 129--137, March 1982.

\bibitem{Aghabozorgi2015}
S.~Aghabozorgi, A.~{Seyed Shirkhorshidi}, and T.~{Ying Wah}, ``{Time-series
  clustering - A decade review},'' \emph{Information Systems}, vol.~53, pp.
  16--38, 2015.

\bibitem{Paparrizos2017-nw}
J.~Paparrizos and L.~Gravano, ``Fast and accurate {Time-Series} clustering,''
  \emph{ACM Transactions on Database Systems (TODS)}, vol.~42, no.~2, p.~8,
  Jun. 2017.

\bibitem{chazal2017introduction}
F.~Chazal and B.~Michel, ``An introduction to topological data analysis:
  fundamental and practical aspects for data scientists,'' 2017.

\bibitem{Turner2019-bc}
K.~Turner and G.~Spreemann, ``Same but different: distance correlations between
  topological summaries,'' Mar. 2019.

\bibitem{Strehl2002ClusterEA}
A.~Strehl and J.~Ghosh, ``Cluster ensembles a knowledge reuse framework for
  combining partitionings,'' in \emph{AAAI/IAAI}, 2002.

\bibitem{1410288}
A.Topchy, M.~H.~C. Law, A.~K. Jain, and A.~L. Fred, ``Analysis of consensus
  partition in cluster ensemble,'' in \emph{Fourth IEEE International
  Conference on Data Mining (ICDM'04)}, Nov 2004, pp. 225--232.

\bibitem{1524981}
A.~Topchy, A.~K. Jain, and W.~Punch, ``Clustering ensembles: models of
  consensus and weak partitions,'' \emph{IEEE Transactions on Pattern Analysis
  and Machine Intelligence}, vol.~27, no.~12, pp. 1866--1881, Dec 2005.

\bibitem{YU201416}
Z.~Yu, L.~Li, H.-S. Wong, J.~You, G.~Han, Y.~Gao, and G.~Yu, ``Probabilistic
  cluster structure ensemble,'' \emph{Information Sciences}, vol. 267, pp. 16
  -- 34, 2014.

\bibitem{10.1007/978-3-319-12640-1_1}
Y.~Wu, X.~Liu, and L.~Guo, ``A new ensemble clustering method based on
  dempster-shafer evidence theory and gaussian mixture modeling,'' in
  \emph{Neural Information Processing}.\hskip 1em plus 0.5em minus 0.4em\relax
  Cham: Springer International Publishing, 2014, pp. 1--8.

\bibitem{shafer1976mathematical}
G.~Shafer, \emph{A Mathematical Theory of Evidence}.\hskip 1em plus 0.5em minus
  0.4em\relax Princeton: Princeton University Press, 1976.

\bibitem{doi:10.1137/0105003}
J.~Munkres, ``Algorithms for the assignment and transportation problems,''
  \emph{Journal of the Society for Industrial and Applied Mathematics}, vol.~5,
  no.~1, pp. 32--38, 1957.

\bibitem{Gitman2018-lb}
I.~Gitman, J.~Chen, E.~Lei, and A.~Dubrawski, ``Novel prediction techniques
  based on clusterwise linear regression,'' Apr. 2018.

\bibitem{Weng2012}
Z.~Weng and X.~Wang, ``Low-rank matrix completion for array signal
  processing,'' in \emph{Acoustics, Speech and Signal Processing (ICASSP), 2012
  IEEE International Conference on}.\hskip 1em plus 0.5em minus 0.4em\relax
  IEEE, 2012, pp. 2697--2700.

\bibitem{Chen2004}
P.~Chen and D.~Suter, ``Recovering the missing components in a large noisy
  low-rank matrix: Application to sfm,'' \emph{IEEE transactions on pattern
  analysis and machine intelligence}, vol.~26, no.~8, pp. 1051--1063, 2004.

\bibitem{nazarovetal2018}
I.~{Nazarov}, B.~{Shirokikh}, M.~{Burkina}, G.~{Fedonin}, and M.~{Panov},
  ``{Sparse Group Inductive Matrix Completion},'' \emph{ArXiv e-prints}, 2018.

\bibitem{yuetal2016}
H.-F. Yu, N.~Rao, and I.~S. Dhillon, ``Temporal regularized matrix
  factorization for high-dimensional time series prediction,'' in
  \emph{Advances in Neural Information Processing Systems 29}, 2016.

\bibitem{Assimakopoulos2000-sy}
V.~Assimakopoulos and K.~Nikolopoulos, ``The theta model: a decomposition
  approach to forecasting,'' \emph{Int. J. Forecast.}, vol.~16, 2000.

\bibitem{Lacombe2018-od}
T.~Lacombe, M.~Cuturi, and S.~Oudot, ``Large scale computation of means and
  clusters for persistence diagrams using optimal transport,'' May 2018.

\end{thebibliography}
